\pdfoutput=1

\documentclass[11pt]{article}

\usepackage{EACL2023}

\usepackage{times}
\usepackage{latexsym}

\usepackage[T1]{fontenc}

\usepackage[utf8]{inputenc}

\usepackage{microtype}

\usepackage{inconsolata}

\usepackage{amsmath}
\usepackage{amsfonts}
\usepackage{tabularx}
\usepackage{hyperref}

\usepackage{bbm}
\usepackage{multicol}
\usepackage{soul}
\usepackage{multirow}
\usepackage{makecell}
\usepackage{caption}
\usepackage{subcaption}
\usepackage{bigdelim}
\usepackage{algorithm}
\usepackage{algpseudocode}
\usepackage{booktabs}
\usepackage{enumitem}
\usepackage{float}
\usepackage{CJKutf8}
\usepackage{hhline}
\usepackage{pifont}
\usepackage{tikz}
\usetikzlibrary{fit, calc}
\usetikzlibrary{positioning}
\usetikzlibrary{shapes,arrows,arrows,positioning,fit}
 \tikzstyle{block} = [rectangle, draw, fill=blue!20, 
 text width=10em, text centered, rounded corners, minimum height=2em]
  \tikzstyle{decision} = [rectangle, draw, fill=red!20, 
 text width=10em, text centered, rounded corners, minimum height=2em]
 \tikzstyle{line} = [draw, -latex']

\newcommand*\circled[1]{\tikz[baseline=(char.base)]{
            \node[shape=circle,draw,inner sep=2pt] (char) {#1};}}
\title{Cascaded Beam Search: Plug-and-Play Terminology-Forcing For Neural Machine Translation}

\author{Fr{\'e}d{\'e}ric Odermatt \and Beni Egressy \and Roger Wattenhofer \\
        ETH Z{\"u}rich, Z{\"u}rich, Switzerland \\
        \texttt{odermafr@ethz.ch  begressy@ethz.ch}}

\begin{document}
\maketitle
\begin{abstract}
This paper presents a plug-and-play approach for translation with terminology constraints. 
Terminology constraints are an important aspect of many modern translation pipelines. 
In both specialized domains 
and newly emerging domains (such as the COVID-19 pandemic), accurate translation of technical terms is crucial.
Recent approaches often train models to copy terminologies from the input into the output sentence by feeding the target terminology along with the input.
But this requires expensive training whenever the underlying language model is changed or the system should specialize to a new domain.
We propose \textbf{Cascade Beam Search}, a plug-and-play terminology-forcing approach that requires no training. 
Cascade Beam Search has two parts: 1) logit manipulation to increase the probability of target terminologies and 2) a cascading beam setup
based on grid beam search,
where beams are grouped by the number of terminologies they contain.
We evaluate the performance of our approach by competing against the top submissions of the WMT21 terminology translation task. 
Our plug-and-play approach performs on par with the winning submissions without using a domain-specific language model and with no additional training.
\end{abstract}

\section{Introduction}

Terminology translation is a key challenge in modern machine translation systems. While most translation systems are trained to be generalists, applications that require accurate translation of terminology are plentiful (e.g., in the bio-medical or legal domains). In addition, new terms and domains can emerge over time, rendering large pretrained language models outdated (e.g., COVID-19). Not only is it difficult and expensive to come by parallel corpora for specialized or emerging domains, but the periodic retraining or fine-tuning of machine translation models can also be energy intensive.

\begin{figure}[H]
    \small\centering
    Source: The \colorbox{blue!15}{WHO} has declared \colorbox{red!20}{COVID-19} a \colorbox{green!20}{pandemic}.\vspace{0.5em}
    
    \underline{French}
    
    {\scriptsize WHO $\rightarrow$ OMS, COVID-19 $\rightarrow$ COVID-19, pandemic $\rightarrow$ pand{\'e}mie}
    \vspace{0.5em}
    
    \small Target: L'\colorbox{blue!15}{OMS} a d{\'e}clar{\'e} que le \colorbox{red!20}{COVID-19} est une \colorbox{green!20}{pand{\'e}mie}.\vspace{0.5em}

    \underline{Korean}
    
    {\scriptsize WHO $\rightarrow$ \begin{CJK}{UTF8}{mj}세계보건기구
    \end{CJK}, COVID-19 $\rightarrow$ \begin{CJK}{UTF8}{mj}코로나19\end{CJK}, pandemic $\rightarrow$ \begin{CJK}{UTF8}{mj}팬데믹\end{CJK}}\vspace{0.5em}
    
        Target: \begin{CJK}{UTF8}{mj}\colorbox{blue!15}{세계보건기구}는 \colorbox{red!20}{코로나19} \colorbox{green!20}{팬데믹을} 선언하였다\end{CJK}.
    \caption{Example of word-level terminology lists and their application in the translation setting.}
\end{figure}

\begin{figure*}[ht]
  \centering
  \includegraphics[width=1.0\linewidth]{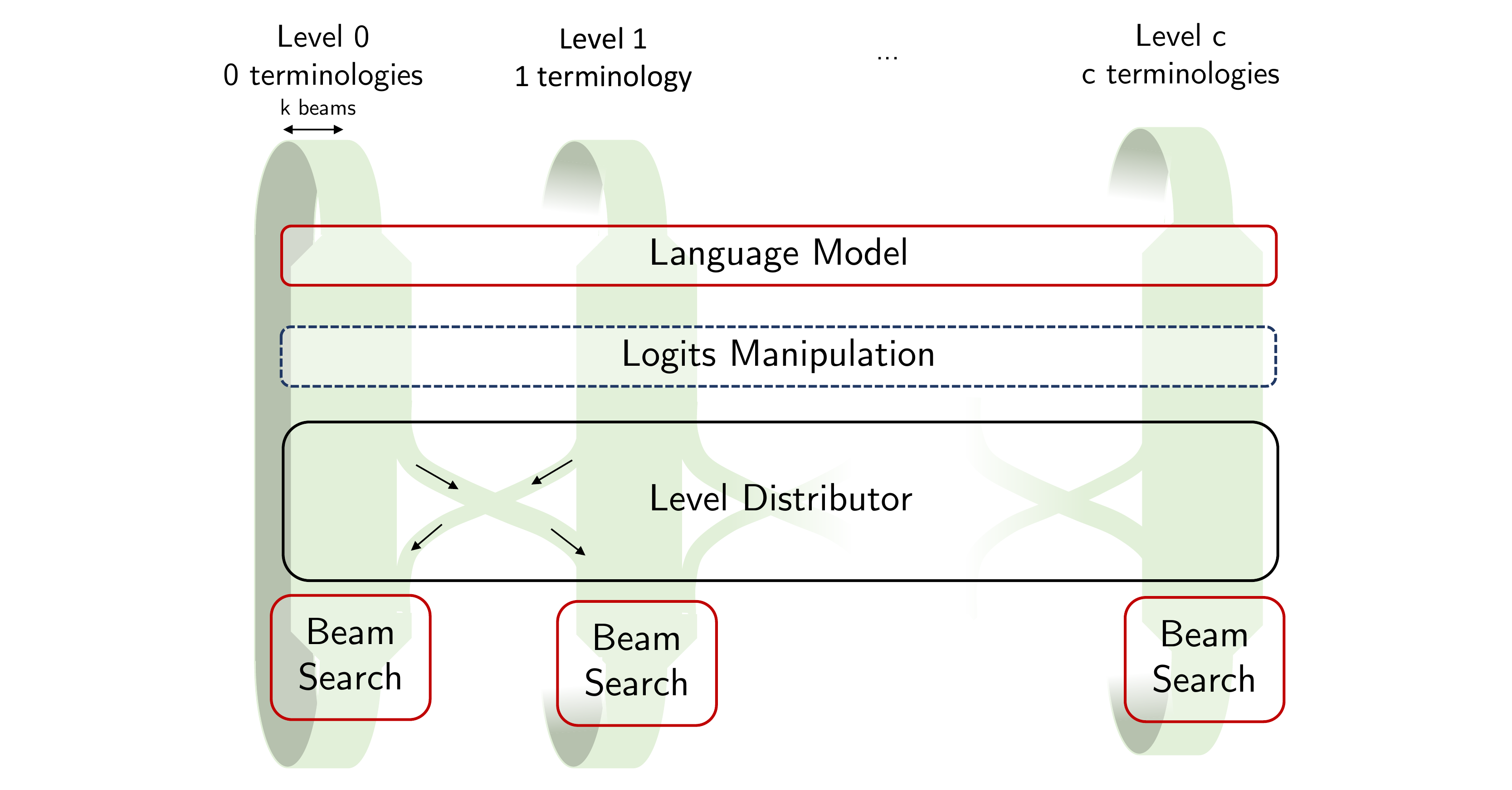}
  \caption{The Cascaded Beam Search setup. Cascaded beam search can be used with any underlying Language Model. It consists of an optional logit manipulation module and a level distributor followed by standard beam search. The logit module increases the probability of desired tokens, the level distributor passes the hypotheses to the correct level and beam search selects the best candidates per level.}
  \label{fig:teaser}
\end{figure*}

In this context, the use of word- or phrase-level terminology lists as also used by human professional translators can be an interesting resource to guide translation. Such lists can be created in a timely manner even for newly emerging domains and if used effectively, they offer the possibility for a flexible neural machine translation (NMT) pipeline that can become an expert in any domain.


NMT with terminology constraints has gained significant attention in recent years culminating in the WMT21 shared task: \emph{Machine Translation using Terminologies} \citep{alam-etal-2021-findings}. Participants were asked to translate COVID-related sentences across five different language pairs\footnote{english-french, english-korean, english-chinese, english-russian, czech-german}.

One of the most successful early approaches to terminology translation comes from \citet{hokamp-liu-2017-lexically} in the form of grid beam search (GBS). In this approach beam search is run on multiple levels in parallel, where each level contains the best beams with a given number of fulfilled constraints. 
This approach can be summarized as trying to place a constraint in every possible position until it finds a good one.
Although very effective, GBS' runtime increases linearly with the number of constraints. To mitigate this issue \citet{post2018fast} propose dynamic beam allocation, which reduces the computational overhead to a constant factor.
More recently, \citet{dinu2019training} have argued that GBS can be brittle under realistic conditions and propose instead to train a language model to copy target terminologies, after first adding these to the input. Indeed all the competitors in the WMT21 competition used some version of this approach.

We argue that it may be time to revisit constrained decoding methods. Our Cascaded Beam Search is based on GBS, but we show that with two important modifications one can achieve significant performance improvements: 1) allowing arbitrary constraint tokenizations and 2) using a cascade level per full constraint instead of per constraint token. These modifications lead to improvements of more than 6 BLEU points across all datasets tested.

To compare against the state-of-the-art in terminology translation, we also evaluate our approach on the WMT21 competition dataset with a generalist underlying multilingual model.
Our decoding approach achieves a near $100\%$ appearance of terminologies, while retaining the BLEU score of the underlying model. This result beats all competitors in terms of terminology appearance, but cannot surpass the BLEU scores of the fine-tuned, competition-winning models as our underlying model 
is limited by the weaker domain-specific translation quality of the underlying model.

\section{Methodology} \label{sec:method}

\subsection{Problem Definition}

A terminology translation task consists of a set $D$ of source and reference target sentence pairs $(s,r)$, and a terminology list $T$ of source and target terminologies. A source terminology can have multiple target translations.



The goal is to translate text such that the output is both a) of high quality and b) incorporates the terminologies correctly. 
We follow \citet{DBLP:journals/corr/abs-2106-11891, alam-etal-2021-findings} and use \texttt{BLEU}\footnote{using sacrebleu with its default tokenizers; inputs are detokenized and true-cased} \cite{papineni-etal-2002-bleu} to evaluate general translation quality, and we use terminology-specific scores to assess the appearance and placement of the terminologies. We give the most important terminology-specific scores below, with $h$ denoting the translation candidate (or hypothesis).
\begin{itemize}[leftmargin=2em]
    \item Exact Match Accuracy (EMA)
    \begin{equation}
        \text{EMA} = \frac{\text{\# matched source terms in } h}{\text{\# source terms}}
    \end{equation}
    
    \item Lemmatized Match Accuracy (LMA) is a lemmatized version of EMA.
    where both the candidate and the target terminologies are lemmatized \citep{qi2020stanza}.
    
    

    
\end{itemize}

\subsection{Beam Search}

Given a language model, the goal of a decoding algorithm is to find the output sequence $\mathbf{y}^*$ that maximizes the conditional probability:
\begin{equation}
    \mathbf{y}^* = \text{argmax}_{\mathbf{y} \in \mathcal{Y}} p(\mathbf{y} \mid \mathbf{x}, \boldsymbol{\theta})
\end{equation}
where $\mathbf{x}$ is the input sentence, $\boldsymbol{\theta}$ are the parameters of the model, and $\mathcal{Y}$ is the set of all sequences in the model vocabulary $V$. 
Most models split the probability up along the decoding time steps, $t$, producing a probability distribution that factors as follows:

\begin{equation}
    p_{\boldsymbol{\theta}}(\mathbf{y}|\mathbf{x}) = \prod_{t=1}^{|\mathbf{y}|} p_{\boldsymbol{\theta}}(y_t | \mathbf{x}, \mathbf{y}_{<t})
\end{equation}

On one hand is exploring all sequences in the exponential room $\mathcal{Y}$ infeasible and on the other hand simple greedy choices at every timestep make locally optimal choices that are likely to result in a globally sub-optimal output. Therefore heuristic approaches such as Beam Search \citep{lowerre1976harpy, sutskever2014sequence} have become the de-facto standard in the machine translation world.

Beam search is a pruned search that keeps a set of $k$ top candidates, looks at all $k \times |V|$ continuations and selects again the top $k$ as candidates for the next time step.

\subsection{Grid Beam Search}
Grid beam search is an extension of beam search that allows for imposing lexical constraints on the output, such as the appearance of specific terms. Given $c$ constraints, grid beam search stores $c+1$ \emph{banks} of candidates, $\{B_i\}_{i=0,\ldots,c}$. Each bank $B_i$ contains the top $k$ candidates that have fulfilled $i$ constraints. 
Although effective, GBS' runtime increases linearly with the number of constraints; moreover the variable total beam size is bad for data parallelism. As a solution to these problems,  \citet{post2018fast} propose dynamic beam allocation, where a fixed total number of beams is distributed among the banks. 
Such decoding methods are also sometimes collectively referred to as constrained beam search. 

Neither of these methods allow for a flexible setup where a constraint may be fulfilled by picking one constraint among a set of possible constraints. For example, one might want to let the system translate \emph{cough} as a noun (\emph{toux}) or a verb (\emph{tousse, tousses, toussons,...}) depending on the context. \citet{li2021guided} introduce an extension of constrained beam search, \textit{disjunctive positive constraints}, that allows for such constraints.
In this paper we will use their decoding algorithm as a baseline, and for simplicity refer to it as grid beam search+ or GBS+, because terminology lists often contain multiple possible target translations for the same source term, as is the case in more recent datasets such as the dataset for the WMT21 shared task on terminology translation.

\subsection{Logit Manipulation} 
\label{subsec:logit}

\citet{pascual-etal-2021-plug-play} introduce a logit modification approach for controlled text generation with keyword constraints. Their approach is aimed at semantically unconstrained text generation such as story writing. In order to encourage certain keywords, they add a vector of cosine similarities to the language model's output distribution so that words similar to the keywords receive a probability boost. 
As well as boosting the probability of keywords, using the cosine similarity can help generate a context in which the keyword can appear more naturally (i.e., with a higher probability score).

In this paper we take a similar approach, but we adapt it to the terminology translation setting. Instead of using cosine similarity for guidance, we use binary encoding to guide only towards the relevant terminologies. 
This makes more sense in our setting, as we would like exact matches in the output and there is enough context in the input sentence to encourage the model to generate a natural context for the desired terminology.

To encourage the terminologies to appear in the output we increase the score of any token that either
\begin{itemize}
    \item continues a target terminology if we are currently producing a target terminology or
    \item starts a target terminology that has not yet appeared if we are not currently producing a target terminology
\end{itemize}



\noindent We modify the logits by adding $\alpha / |T'|$ to the desired tokens $T'$, and then re-normalize with softmax
to get a probability distribution: 
\begin{equation*}
    \text{probs} = \text{softmax}\Big(\text{softmax}(x) + \alpha \cdot \frac{\mathbbm{1}_{T'}}{|T'|}\Big),
\end{equation*}
\noindent where $x$ is the vector of logits from the language model, $\mathbbm{1}$ is the indicator function, and $\alpha \in \mathbb{R}$ is a constant that can be tuned.
The approach is flexible with regards to the tokens that are included in $T'$. Tokens that start or continue terminologies can be added to $T'$ to encourage their appearance. 
More analysis on logit manipulation and choices of tokens in $T'$ can be found in Section~\ref{sec:experiments_logman}. 

\subsection{Cascaded Beam Search}

This paper proposes an inference-time only modified constrained beam search variant that can enforce terminology appearance during a translation task 
by keeping different sets of beams (so-called banks) for different levels of progress in generating the target terminologies. As visible in Figure \ref{fig:teaser}, cascaded beam search is made up of three parts: the language model, a cascade level distributor and finally a classical beam search procedure per level. It can also be combined with an optional logit manipulation module.\vspace{0.5em}

\noindent\textbf{Language Model}: The proposed method can be applied in a plug-and-play manner to any autoregressive machine translation model that outputs probabilities over a fixed vocabulary.\vspace{0.5em}

\noindent\textbf{Logit Manipulation (optional)}: Cascade beam search can optionally be combined with logit modification as described in Section \ref{subsec:logit}. \vspace{0.5em}

\noindent\textbf{Cascade Level Distributor}: The level distributor decides which level each hypothesis should be allocated to based on the next token. The n\textsuperscript{th} level contains hypotheses with n terminologies, either with n complete terminologies or part-way through the n\textsuperscript{th} terminology. Unlike GBS, we consider any token whose characters begin or continue a terminology to be successful.  
The Cascade Level Distributor is described in more detail in Figure \ref{fig:level_distributor} in the appendix. Differences to GBS are described in detail in Section \ref{subsec:differences}. \vspace{0.5em}

\noindent\textbf{Beam search per cascade level}: After all continuations are distributed among the cascade levels we apply a standard beam search step per level: We pick the top $k$ ($k$: number of beams) hypotheses that don't end with an end of sequence token, \texttt{<EOS>}. 
We move any top-scoring hypotheses that do end with \texttt{<EOS>} at the highest level to a list of final hypotheses. 
We stop when we have $k$ hypotheses that fulfill all the constraints in our list of final hypotheses, or when we reach the maximum sequence length. We then pick the highest scoring sequence amongst the final hypotheses fulfilling the most constraints as our output sequence. \vspace{0.5em}

\noindent\textbf{Complexity}: Since cascaded beam search requires one set of beams per cascade level, the total number of hypotheses grows linearly with the number of terminologies. The total number of beams is $k(c+1)$. In practice the number of terminologies should be limited or dynamic beam allocation \citep{post2018fast} should be used. For simplicity and better comparability we opt for using the grid beam search setup in our experiments, and we leave runtime optimization through dynamic beam allocation to future work.
\vspace{0.5em}

Our approach is implemented within the \texttt{transformers} library's \texttt{generate} function and can be flexibly applied to any language model that is ported to this function. The code will be made publicly available at the time of publication.

\subsection{Differences to Grid Beam Search}
\label{subsec:differences}


Cascaded beam search has two key differences to grid beam search.
Firstly, GBS uses a fixed tokenization of the terminology constraints. This restricts constraint fulfillment to generating the exact sequence of tokens produced by the tokenizer. In contrast to GBS, cascaded beam search allows for any possible tokenization of a terminology constraint. We achieve this by checking whether the alphanumeric characters that make up the token are a successful start or continuation of a terminology. 
We have considered the qualities that this approach brings. Of course, checking all the tokens that constitute alphanumeric continuations of a terminology carries additional computational burden. However, this can be implemented efficiently using Trie-based vocabulary representations. On the other hand, we argue that requiring a fixed tokenization of a lexical constraint can be a quality deprecating factor for multilingual language models that operate on huge vocabularies using subword tokenizers, such as WordPiece \citep{wordpiece} or SentencePiece \citep{sentencepiece}, where words can have a large amount of valid tokenizations.

Secondly, we use a single level per terminology constraint, rather than a new level per target terminology token. This is better aligned with the aim of the task and the appearance-based evaluation metrics, where each terminology constraint receives the same weight in the metric regardless of terminology token length. 
In addition this also reduces the total number of beams and therefore the runtime and memory requirements of our decoding algorithm.

\section{Experimental Setup} \label{sec:setup}
\subsection{Datasets}

First we compare cascaded beam search against an existing constrained beam search method \citep{li2021guided} on the WMT17 German-English news translation task\footnote{\href{https://www.statmt.org/wmt17/translation-task.html}{https://www.statmt.org/wmt17/translation-task.html}}. The terminology sets, $T$, are taken from \citet{dinu2019training}. They were extracted them from Wiktionary and IATE to create two versions of the terminology translation task.

\begin{table}[h!]
    \centering
    \resizebox{\linewidth}{!}{%
    \begin{tabular}{c|c|c|ccccc}
        \toprule
        Language & \multirow{2}{*}{Dataset} & \multirow{2}{*}{Size} & \multicolumn{5}{c}{Number of Terminologies}\\
        Pair & &  & 0 & 1 & 2 & >=3 & max\\
        \midrule
        \multirow{2}{*}{EN-DE} & WIKT & 727 & 0.0\% & 81.7\% & 15.3\% & 2.7\% & 4 \\
        & IATE & 414 & 0.0\% & 91.3\% & 8.2\% & 0.5\% & 3 \\
        \bottomrule
    \end{tabular}
    }
    \caption{Overview of the WMT17 data split that contains IATE or WIKT terminologies. max refers to the maximum amount of annotated terminologies in a single sample.}
    \label{tab:overview_wmt17_data}
\end{table}

We also evaluate our approach on the WMT21 shared task on terminology translation \citep{alam-etal-2021-findings}. As our method requires no retraining we evaluate our approach on the test set splits that were used for the competition. The dataset covers the bio-medical domain with a specific focus on new terminology that appeared during the COVID-19 pandemic.
The dataset is annotated on the source side with possible terminology translations if a target terminology is found in the reference target translation. Note that considerable parts of the dataset do not contain terminologies at all.

\begin{table}[h!]
    \centering
    \resizebox{\linewidth}{!}{%
    \begin{tabular}{c|c|ccccc}
        \toprule
        Language & \multirow{2}{*}{Size} & \multicolumn{5}{c}{Number of Terminologies}\\
        Pair & & 0 & 1 & 2 & >=3 & max\\
        \midrule
        EN-FR & 2100 & 40.2\% & 30.0\% & 15.4\% & 14.4\% & 11 \\
        EN-KO & 2100 & 66.9\% & 22.6\% & 6.6\% & 3.9\% & 9 \\
        EN-RU & 2100 & 44.3\% & 30.1\% & 15.4\% & 10.2\% & 9\\
        CZ-DE & 3426 & 6.5\% & 61.2\% & 25.8\% & 6.5\% & 6\\
        \bottomrule
    \end{tabular}
    }
    \caption{Overview of the test set splits for the WMT21 shared task on terminology translation dataset.}
    \label{tab:my_label}
\end{table}

\subsection{Base Models} \label{sec:basemodels}

As our decoding algorithm can be used in a plug-and-play fashion with any neural machine translation model with an autoregressive decoder, we evaluate the performance of cascaded beam search on a set of base models.\smallskip

\begin{figure*}[ht]
    \centering
    \includegraphics[width=\linewidth]{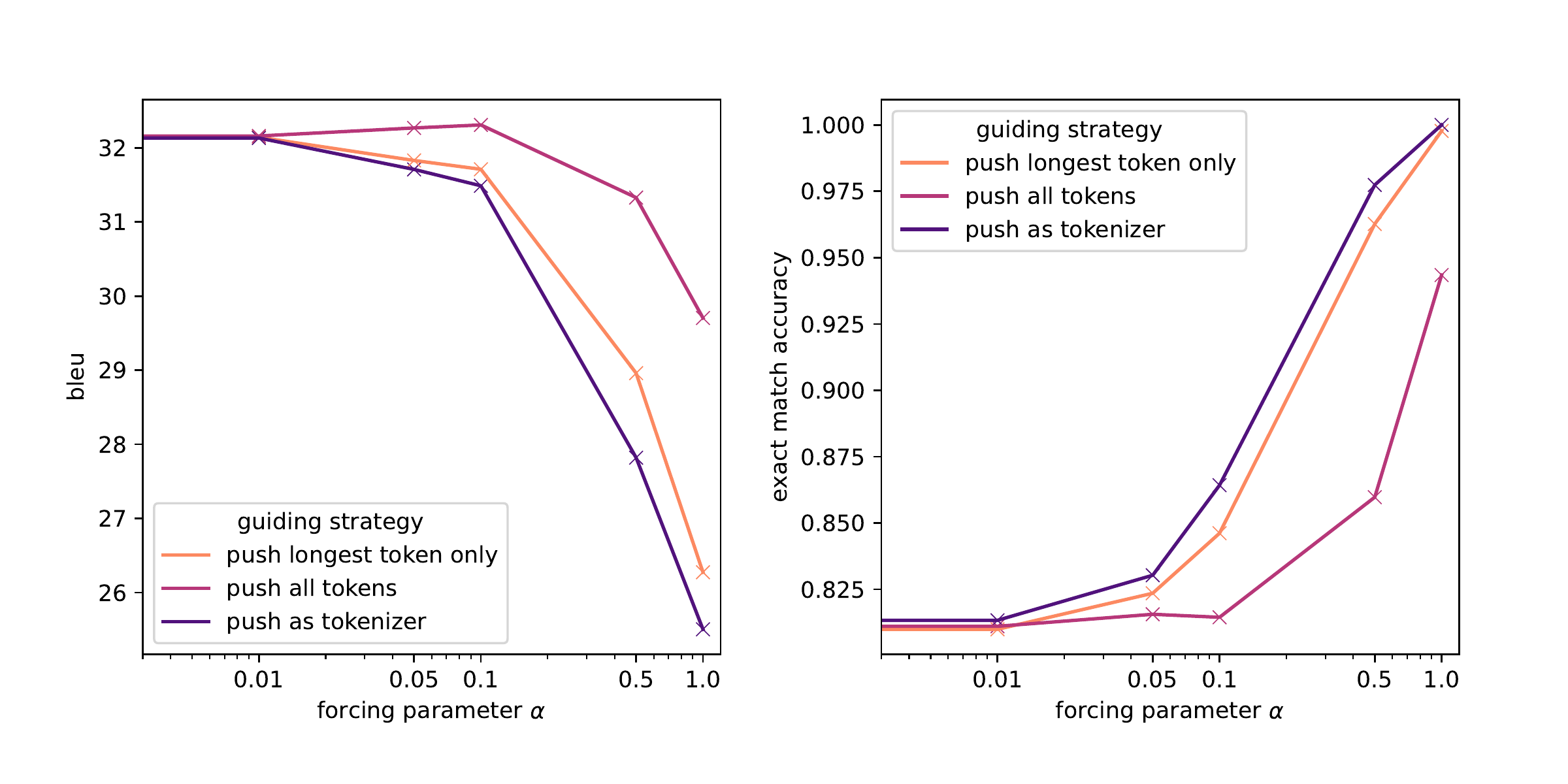}
    \vspace{-0.9cm}
    \caption{Comparison of BLEU and exact match accuracy (EMA) scores for different logit modification scenarios, when using logit modification
    without cascaded beam search. \emph{Push longest} increases the probability of only the longest token that begins/continues a terminology, \emph{push all} increases the probability for all tokens that begin/continue a terminology and \emph{push tokenizer} increases the probability only for tokens produced by the tokenizer. All results are on the WMT17 dataset with the Wiktionary based terminology set.}
    \label{fig:WIKT_alpha}
\end{figure*}

\noindent\textbf{M2M100}: M2M100 \citep{m2m100} is a many-to-many multilingual translation model that works across 100 languages. Sentences are tokenized using SentencePiece \citep{sentencepiece}, the model is a transformer based seq-to-seq model made up of an encoder and decoder module. The model is available in two different sizes, M2M100 small with 418M parameters and M2M100 large with 1.2B parameters. Of particular relevance to the WMT21 competition task, is that this model and the associated paper were released in late summer 2020. As a result, the model was exposed to little, if any, training data from after the COVID-19 outbreak. We therefore use M2M100 as a \emph{pre-COVID} reference model. This makes the WMT21 task, which is focused on translating COVID related texts, a particularly challenging and realistic scenario for terminology translation with this model. One simple confirmation that this is the case, is that the word COVID-19 is tokenized as [CO,V,ID,-,19] by M2M100's sentencepiece tokenizer, which is a comparably high number of tokens for a word that should appear frequently in data after the start of the pandemic.\smallskip

\noindent\textbf{NLLB}: "No Language is Left Behind" \citet{nllb} was released in July 2022 and can be seen as an extension of the M2M100 project. NLLB covers 200 languages, many of which are considered to be "low resource" languages. The translation model is a sparsely-activated MoE (Mixture of Experts) transformer-based model. 
As the dataset was mined more recently, it includes data from after the COVID outbreak, so we treat this model as a \emph{post-COVID} baseline for the WMT21 task. The model also outperforms M2M100 on various translation benchmarks, which should be taken into consideration when looking at the results. Interestingly, NLLB has reverted back to basic sampling methods instead of using beam search, and we report scores attained as such in Tables~\ref{tab:wmt21_short}~and~\ref{tab:wmt21_full}. 

\section{Experiments} \label{sec:results}

\subsection{Logit Modification}
\label{sec:experiments_logman}

\begin{table*}[!h]
    \centering
    \resizebox{\linewidth}{!}{%
    \begin{tabular}{l|l|l|ccc|l|ccc}
        \toprule[1.5pt]
        Model & Decoding Method & & EMA & LMA & BLEU & & EMA & LMA & BLEU \\
        \midrule
        \multirow{5}{*}{M2M100 large} & baseline & \multirow{5}{*}{WIKT} & 0.818 & 0.858 & \textbf{32.53} & \multirow{5}{*}{IATE} & 0.810 & 0.854 & \textbf{32.43} \\
        & logit only push longest ($\alpha = 0.1$) & & 0.846 & 0.883 & 31.71         & & 0.836 & 0.878 & 31.32\\
        & grid beam search & & 0.977 & \textbf{1.000} & 24.99 & & 0.984 & 0.998 & 23.93\\
        & grid beam search + & & 0.977 & \textbf{1.000} & 24.99 & & 0.984 & 0.998 & 23.93\\
        & cascaded & & \textbf{1.000} & \textbf{1.000} & 31.68 & & \textbf{1.000} & \textbf{1.000} & 31.24\\

        \bottomrule[1.5pt]
    \end{tabular}
    }
    \caption{Comparison of translation quality on the WMT17 datasets using different decoding methods. The baseline is standard beam search, grid beam search + refers to the extended decoding algorithm taken from \citet{li2021guided} and cascaded beam search is the method we propose. The highest scores are highlighted in bold.}
    \label{tab:wmt17}
\end{table*}

Logit modification is a very flexible approach. There are many possible options for the set of guide tokens $T'$, and the strength of forcing parameter $\alpha$, can also be varied. We compare three different options for $T'$:
\begin{itemize}
    \item \textbf{push tokenizer} manipulates the tokens produced by the tokenizer (in order)
    \item \textbf{push longest} manipulates the \emph{longest} token that begins or continues a terminology
    \item \textbf{push all} manipulates \emph{all} tokens that begin or continue a terminology
\end{itemize}

Logit modification can be used without cascaded beam search as a decoding method on its own. 
We compare the above options under this setting for analysis. 
Figure \ref{fig:WIKT_alpha} shows the BLEU and EMA scores on the Wiktionary dataset from WMT17 as the forcing strength is varied. The corresponding results for the IATE dataset from WMT17 can be seen in Figure \ref{fig:IATE_alpha} in the appendix.

\subsection{Cascaded Beam Search with Logit Modification}
\label{sec:experiments_cascaded_logman}

\begin{figure*}[h!]
    \centering
    \includegraphics[width=\linewidth]{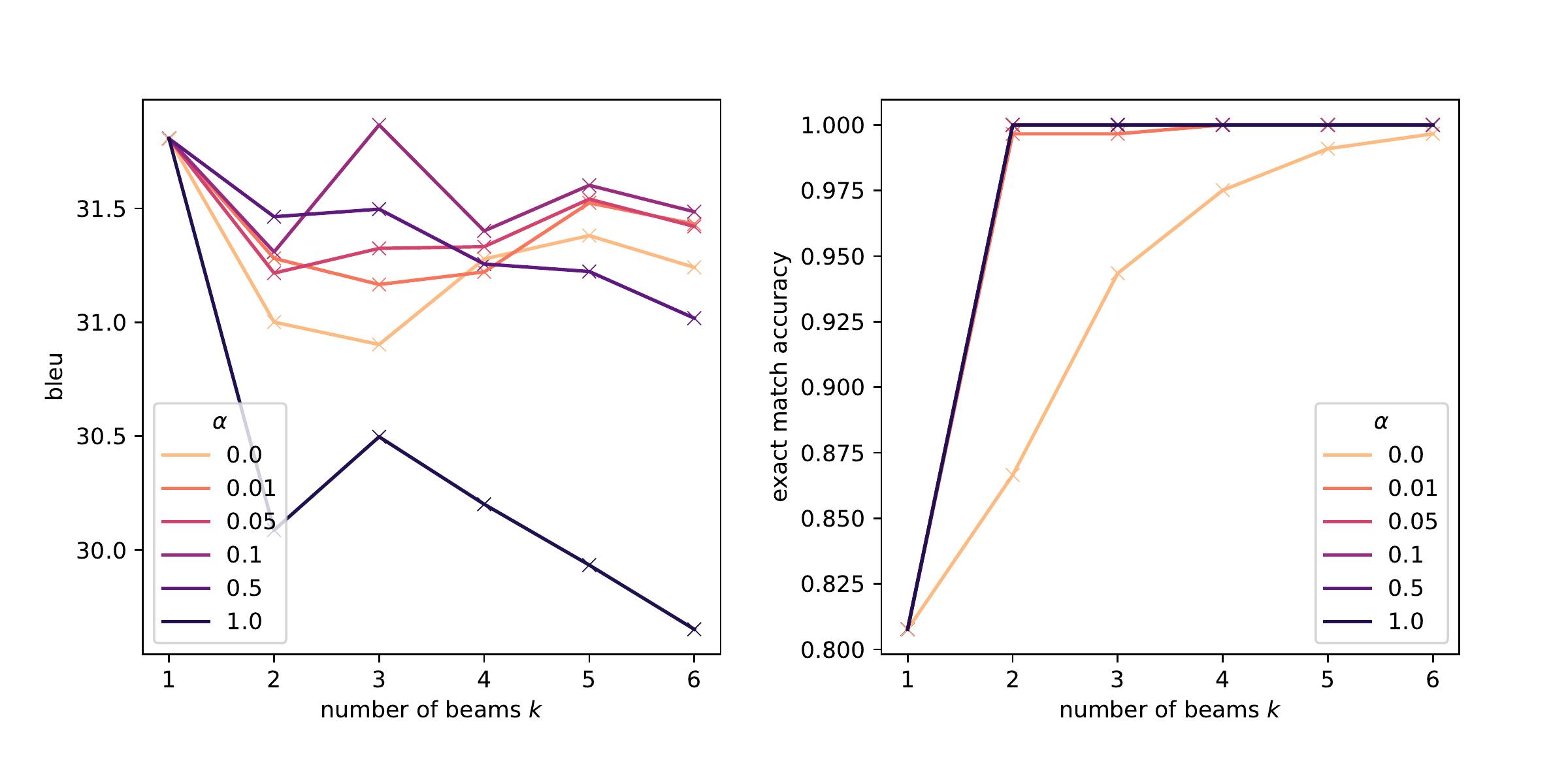}
    \vspace{-0.9cm}
    \caption{Comparison of BLEU and exact match accuracy (EMA) scores under different settings when using cascaded beam search (CBS) with logit modification. We vary the strength of the logit modification, $\alpha$, and the beam sizes per cascade level, $k$. Note that $\alpha=0$ corresponds to CBS with no logit modification. All results are on the WMT17 dataset with the Wiktionary based terminology set.}
    \label{fig:WIKT_beams_x_alpha}
\end{figure*}

We now look at the combination of cascaded beam search with logit modification. In particular we analyse when and how much logit modification helps when applied on top of cascaded beam search.
Figure \ref{fig:WIKT_beams_x_alpha} shows how the BLEU score and terminology appearance (EMA) change as we vary the strength of the logit modification, $\alpha$, for different beam sizes per cascade level.

\subsection{Comparison to Constrained Beam Search}
\label{sec:experiments_wmt17}

Using the WMT17 dataset, we compare cascaded beam search with existing constrained beam search implementations: grid beam search (GBS) \citep{hokamp-liu-2017-lexically}, and a modification of GBS that allows for multiple targets for a single source terminology (GBS+) \citep{li2021guided}. We also include a baseline using standard beam search. For a fair comparison, we use the same underlying translation model, namely M2M100large \citep{m2m100}, and the same beam size ($k=5$) for all of the constrained decoding algorithms and we use $5(c+1)$\footnote{This corresponds to the total number of beams used by the constrained decoding methods, $k(c+1)$.} beams for beam search.
Cascade beam search is applied with logit modification with a guidance strength of $\alpha=0.2$.
We also include logit modification only results for comparison.
The results are shown in Table \ref{tab:wmt17}. 

\subsection{Comparison to SOTA models on WMT21} 
\label{sec:experiments_wmt21}

We compare Cascaded Beam Search to the winning submissions of the WMT21 shared task on terminology translation: PROMT \citep{promt} for the English-French and English-Russian datasets, and Kakao Enterprises (KEP) \citep{bak-etal-2021-kakao} for the English-Korean and Czech-German datasets. We include M2M100 using standard beam search, to give a baseline for the general translation quality of this base model. Note that the winning submissions used translation models fine-tuned for the bio-medical setting of the task, so we do not expect our model to outperform these submissions in BLEU scores. Therefore the M2M100 with standard beam search serves as a much needed baseline. 

Finally we include NLLB as a post-COVID baseline. M2M100 was specifically chosen as a pre-COVID model that would need help with the terminologies, but we also wanted to see how a more recent translation model would perform without any terminology guidance.

The results for the English-French and English-Russian datasets can be seen in Table \ref{tab:wmt21_short}. A full set of results can be found in Table \ref{tab:wmt21_full} in the appendix.

\begin{table}[H]
\centering

\caption{Comparison of translation quality between the WMT21 competition winners and our system. The highest scores are highlighted in bold.}
\label{tab:wmt21_short}
{\normalsize
\resizebox{\linewidth}{!}{%
\begin{tabular}{ll|cc|c}
\toprule
\multirow{1}{*}{}         & \multirow{1}{*}{System} & EMA               & LMA               & BLEU  \\\midrule
\multirow{8}{*}{EN-FR} & PROMT.soft                 & 0.959             & 0.972             & 41.22 \\
                            & M2M100 small baseline & 0.813             & 0.838             & 35.38 \\
                            & M2M100 small cascaded & 0.985             & 0.990             & 35.21\\
                            & M2M100 small GBS+      & 0.989             & 0.993             & 21.86 \\
                            & M2M100 small DBA/GBS new & 0.983  & 0.993 & 27.92 \\
                            & M2M100 large baseline & 0.880             & 0.899             & 40.15 \\
                            & M2M100 large cascaded & \textbf{0.991}    & \textbf{0.994}    & 40.51\\
                            & M2M100 large GBS+      & 0.994             & 0.996             & 27.76 \\
                            & M2M100 large DBA/GBS new & 0.994 & 0.997 & 35.53 \\
                            \cmidrule{2-5}
                            & NLLB                  & 0.903             & 0.922             & \textbf{46.68}\\

\midrule
\multirow{8}{*}{EN-RU} & PROMT.soft                 & 0.862             & 0.913             & \textbf{31.22} \\
                            & M2M100 small baseline & 0.682             & 0.725             & 23.89 \\
                            & M2M100 small cascaded & 0.916             & 0.934             & 23.73\\
                            & M2M100 small GBS+      & 0.963             & 0.969             & 12.01 \\
                            & M2M100 small DBA/GBS new & 0.958 & 0.967 & 18.85 \\
                            & M2M100 large baseline & 0.753             & 0.797             & 29.09 \\
                            & M2M100 large cascaded & \textbf{0.925}    & \textbf{0.950}    & 29.22\\
                            & M2M100 large GBS+      & 0.973             & 0.982             & 16.56 \\
                            & M2M100 large DBA/GBA new & 0.970 & 0.981 & 27.67 \\
                            \cmidrule{2-5}
                            & NLLB                  & 0.782             & 0.827             & 27.76\\

\bottomrule
\end{tabular}
}
}
\end{table}

\section{Discussion} \label{sec:discussion}
\subsection{Logit Modification}
\label{sec:discussion_logman}

Figure \ref{fig:WIKT_alpha} shows the results on the WMT17 Wiktionary dataset under different logit modification settings.
Firstly, we can see on the right that logit modification has the desired effect: As we increase the forcing parameter $\alpha$, the appearance rate of the terminologies goes up, even reaching $100\%$. However on the left we can see that this comes at the cost of a decreasing BLEU score.
Despite the boost to BLEU from the appearance of the terminologies, the score still drops by about $5$ points.
There is also no clear sweet spot, the BLEU score decreases just as the EMA increases.
\emph{Push longest} and \emph{push tokenizer} behave very similarly, but push all seems to lag behind both in increasing EMA and in decreasing BLEU. This makes sense intuitively; in the case of \emph{push all} the forcing is spread over a much larger set of token so each token in $T'$ receives less of a push overall.

\subsection{Cascaded Beam Search with Logit Modification}
\label{sec:discussion_cascaded_logman}

Figure \ref{fig:WIKT_beams_x_alpha} shows how cascaded beam search with logit modification performs on the WMT17 Wiktionary dataset. 
Logit modification has the greatest effect on EMA for small values of $k$. Even with a very low $\alpha$, we achieve almost $100\%$ EMA when using only $k=2$ beams. In contrast, without logit modification $6$ beams are required to achieve a similar EMA score. 
In the left plot we see that especially for smaller values of $k$ there is also an increase in the BLEU score when using logit modification. This higher BLEU score will partly be due to the higher appearance of the terminologies in the output, and not necessarily due to better general translation quality. 
However, increasing the parameter too far (e.g., $\alpha=1.0$) severely damages the BLEU score, especially for larger beam sizes. 
We show similar results for the IATE dataset from WMT17 in the appendix.
Looking at both sets of results, we see that a beam size of around $3$ to $5$ and a guidance parameter of $\alpha=0.1$ gives the best combined scores for these datasets.

These results assure us that the combination of these two decoding approaches can give the best terminology translation results, especially when a restricted computational budget is available.

\subsection{Comparison to Constrained Beam Search}
\label{sec:discussion_wmt17}

Table \ref{tab:wmt17} shows a comparison of cascaded beam search with GBS and a modified version referred to as GBS+. 
Note that since this dataset does not have multiple target translations for any terminology, the results for GBS and GBS+ are identical. 
The beam search baseline achieves an exact terminology appearance rate of around $81\%$ and a BLEU score of around $32.5$ on the datasets.
We can see that GBS reaches an almost perfect EMA score on both datasets, but looses around $7$ BLEU points compared to the baseline. On the other hand cascaded beam search reaches exactly a $100\%$ appearance rate, and loses only around $1$ BLEU point on the datasets, clearly outperforming GBS.

\subsection{Comparison to SOTA models on WMT21} 
\label{sec:discussion_wmt21}

The BLEU score of the M2M100 baselines is significantly lower for most language pairs when compared to the competition winning models. This is to be expected given that M2M100 has not been tuned to the task domain. For the EN-FR language pair NLLB reaches the impressive BLEU score of $46.68$, but otherwise the competition winners have the highest BLEU scores overall. On the other hand, the exact match accuracy (EMA) and lemmatized match accuracy (LMA) are already relatively high for the baselines, but rise to values close to $100\%$ for Cascaded Beam Search. Indeed cascaded beam search attains the highest EMA and LMA scores in all language pairs. 

What is again striking, is that cascade beam search clearly outperforms GBS in terms of BLEU whilst also attaining higher terminology appearance rates. Moreover even when compared to the M2M100 baselines, a trend for slightly increased BLEU scores can be seen when using cascaded beam search.

\section{Conclusion}
\label{sec:futurework}



We introduce two decoding methods for terminology translation: cascade beam search and logit modification. Both methods can improve terminology appearance on terminology translation datasets. However we show how they can be combined to clearly outperform existing decoding approaches.
Cascade beam search with logit modification can be combined with a pre-trained generalist multilingual model and still achieve competitive results on domain-specific tasks. This makes cascade beam search a very versatile alternative to the current state-of-the-art methods that require fine-tuning.
Finally, the analysis shows that even with very small beam sizes, logit modification helps cascade beam search attain its potential with lower computational requirements.

We feel that inference-time only decoder modifications, like cascaded beam search or grid beam search, might be overlooked by current research in terminology translation and can be used to greater effect.



\section*{Limitations} \label{sec:limitations}
Whilst we have given several reason that could explain why cascaded beam search performs significantly better than standard grid beam search, a detailed analysis and ablation study would help further our understanding. 

In addition, like grid beam search, cascaded beam search also has an unfavourable runtime that grows linearly in the number of constraints. Although we restricted ourselves to this setting for simplicity and comparability, dynamic beam allocation should be added to improve the runtime. We leave this analysis to future work. 


\section*{Ethics Statement}
We understand that terminology translation can be used for distributing misinformation more widely or generating harmful content. However, we believe that further research into automatic translation can equip us with the necessary tools to identify, correct and prevent malicious use of these methods.
Moreover, terminology translation is a valuable tool for ensuring accurate translations and reducing inadvertent misinformation. Correct translations are especially important in specialized and critical domains, such as, translating COVID advice or heavy machinery instructions.
Successful methods can also be a valuable asset for low resource languages, where terminology dictionaries are much easier to come by than large amounts of domain-specific training data.
Finally, we consider the environmental impacts of our method. As with grid beam search, the approach can be runtime intensive, and we encourage further work to reduce this computational overhead. However on the positive side, the method is completely plug-and-play. This means it is able to make use of pre-trained language models that require extensive training. This can significantly reduce the energy requirements for setting up a terminology translation system.

\bibliography{anthology,custom}
\bibliographystyle{acl_natbib}

\clearpage

\appendix

\onecolumn
\section{Cascade Level Distributor}
\label{sec:app_cascade_figure}

\begin{figure}[H]

\begin{subfigure}[b]{\textwidth}
\centering
\scalebox{0.9}{

 	\begin{tikzpicture}[node distance = 2cm, auto]\label{ams1}
 	\node [block] (init) {current hypothesis \\ \& new token};
  	\node [decision, below=1.5cm of init] (partof) {last token was part of terminology?};
 	\node [decision, below right=1.5cm and -0.8cm of partof] (finisher) {last token finished terminology we were producing?};
 	\node [decision, below left=1.5cm and -0.4cm of partof] (inactive) {next token starts new terminology?};

 	\node [decision, below left=1.5cm and -0.8cm of finisher] (snt) {new token starts new terminology?};
 	\node [decision, below right=1.5cm and -0.8cm of finisher] (continues) {next token continues terminology?};
 	\node [decision, below right=1.5cm and -3.0cm of continues] (restart) {next token restarts other terminology?};

 	\node [decision, below of= snt] (nottermbutnewword) {new token is a new word?};

 	\node [block, below left=12cm and 1cm of init] (levelup) {go up level};
 	\node [block, below right=12cm and 1cm of init] (leveldown) {go down level};
 	\node [block, below=12cm of init] (staylevel) {stay on level};

 	\node [coordinate, below=6cm of inactive] (inactivecoord) {};
 	\path [line] (init) -- (partof);
 	\path [line] (partof) -| node {yes} (finisher);
 	\path [line] (partof) -| node[left] {no} (inactive);

 	\path [line] (finisher) -| node[left] {yes} (snt);
 	\path [line] (finisher) -| node {no} (continues);

 	\path [line] (snt) -| node[left] {\circled{1} yes} (levelup);
 	\path [line] (snt) -- node {no} (nottermbutnewword);
 	\path [line] (continues) -- node {no} (restart);

  	\path [line] (nottermbutnewword) -- node[left, pos=0.3] {\circled{4} yes} (staylevel);
  	\path [line] (continues) -- node[left, pos=0.3] {\circled{5} yes} (staylevel);
  	\path [line] (restart) -- node[above, pos=0.2] {yes \circled{6}} (staylevel);
  	\path [line] (restart) -- node[right] {no \circled{8}} (leveldown);
    \path [draw,->] (inactive) -- (inactivecoord) -- node[left, pos=0.0]{yes \circled{2}} (levelup);
  	\path [line] (inactivecoord) -- node[above, pos=0.5] {no \circled{3}} (staylevel);
  	\path [line] (nottermbutnewword) -- node[below, pos=0.1] {no \circled{7}} (leveldown);

 	\end{tikzpicture}
 	}
 	\caption{Decision Tree for the Cascade Level Distributor. The Cascade Level Distributor considers all possible hypothesis and new token combinations and delegates them to the appropriate cascade level.}
\end{subfigure}%
\vspace{0.5cm}
\begin{subfigure}[b]{\textwidth}
    Examples: Force terminologies ``COVID-19'' and ``SARS-COV-2'', new token in examples underlined.\medskip\\
    \circled{1} The last token was part of a terminology, but it was also its last token and the new token starts a new terminology. Example: ``COVID-19 \underline{SARS}''  
    \smallskip
    \\
    \circled{2} We were not in a terminology but the next token starts a terminology. Example: ``\ldots but is \underline{COV}''
    \smallskip
    \\
    \circled{3} We were not in a terminology and the next token does not start one. Example: ``\ldots but is \underline{coffee}'' 
    \smallskip
    \\
    \circled{4} The last token was part of a terminology, but it was also its last token. The new token is a new word, but not the start of a terminology. Example: ``\ldots but COVID\underline{ is}'' 
    \smallskip
    \\
    \circled{5} The last token was part of a terminology, but it was not the final token. The new token continues an ongoing terminology. Example: ``\ldots COV\underline{ID}'' 
    \smallskip
    \\
    \circled{6} The last token was part of a terminology, but it was not the final token. The new token is not one that continues the active terminology, but it starts another terminology. Example: ``\ldots COV\underline{ SAR}'' 
    \smallskip
    \\
    \circled{7} The last token was part of a terminology, but it was also its last token. The new token is not the start of a word or punctuation though. Example: ``\ldots COVID-19\underline{90}''
    \smallskip
    \\
    \circled{8} The last token was part of a terminology, but it was not the final token. The new token does not continue the terminology and is also not the start of another terminology. Example: ``\ldots COV\underline{ERT}''
\end{subfigure}%
\caption{The Cascade Level Distributors decision logic and examples}
\label{fig:level_distributor}
\end{figure}

\twocolumn

\section{WMT17 IATE results}
\label{sec:appendix}

\vfill
\hspace{2em}

\begin{figure}[!htb]
    \centering
    \includegraphics[width=\linewidth]{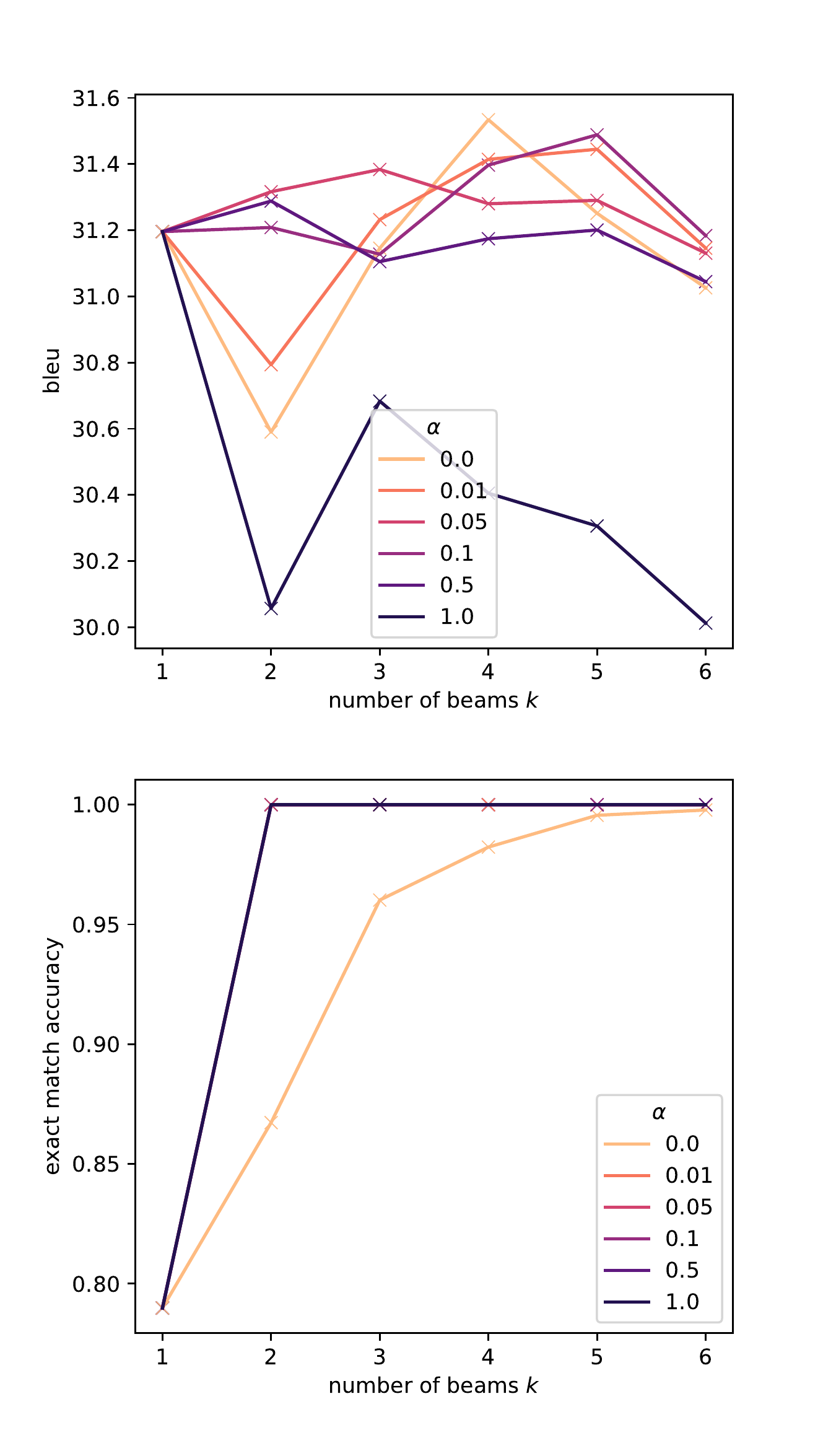}
    \caption{Comparison of BLEU and exact match accuracy (EMA) scores under different settings when using cascaded beam search (CBS) with logit modification. We vary the strength of the logit modification, $\alpha$, and the beam sizes per cascade level, $k$. Note that $\alpha=0$ corresponds to CBS with no logit modification. All results are on the WMT17 dataset with the IATE based terminology set.}
    \label{fig:IATEnumbeam}
\end{figure}

\vfill
\begin{figure}[!htb]
    \centering
    \includegraphics[width=\linewidth]{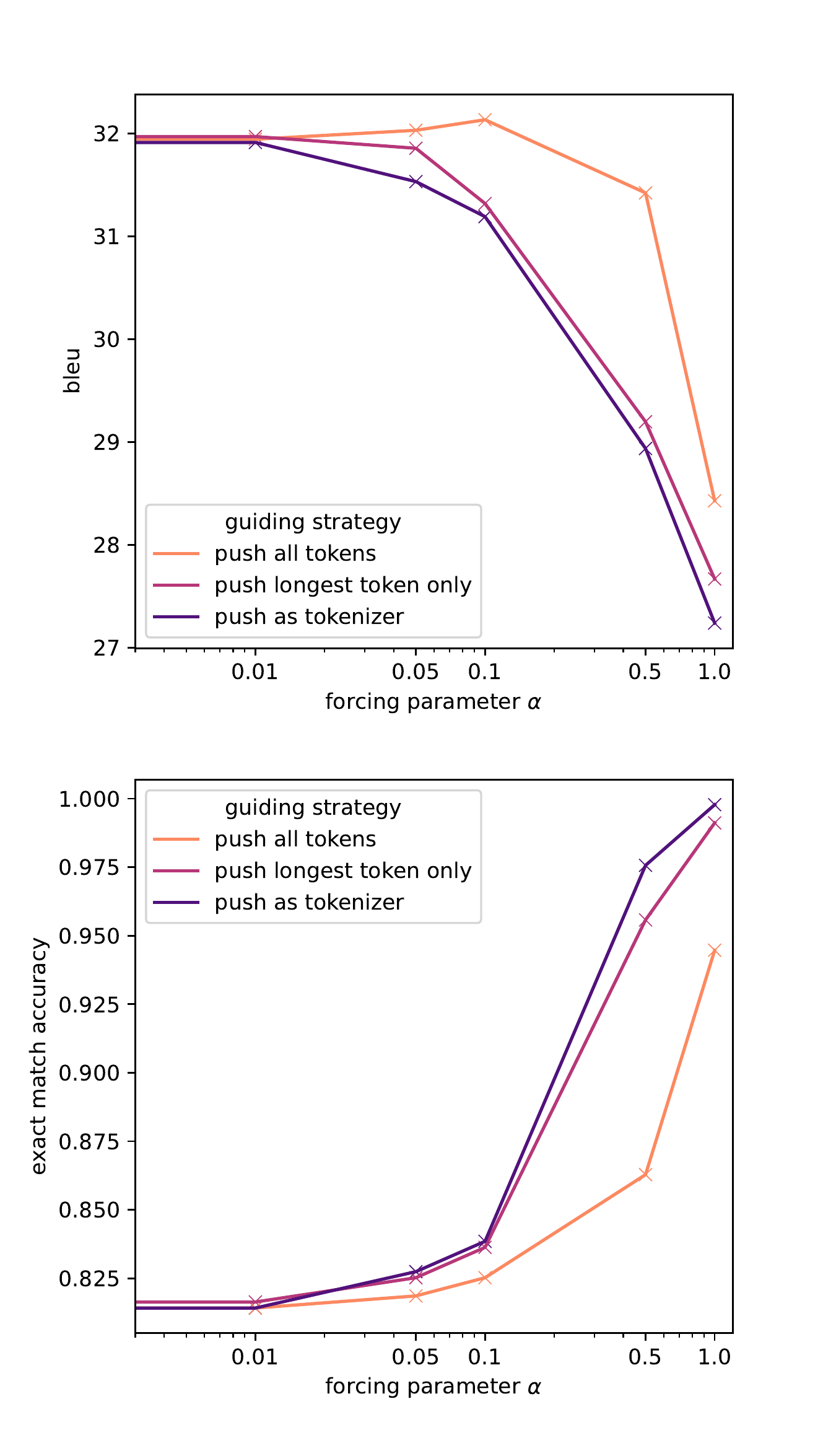}
    \caption{Comparison of BLEU and exact match accuracy (EMA) scores for different logit modification scenarios, when using logit modification
    without cascaded beam search. \emph{Push longest} increases the probability of only the longest token that begins/continues a terminology, \emph{push all} increases the probability for all tokens that begin/continue a terminology and \emph{push tokenizer} increases the probability only for tokens produced by the tokenizer. All results are on the WMT17 dataset with the IATE based terminology set.}
    \label{fig:IATE_alpha}
\end{figure}

\vfill

\newpage\clearpage

\begin{table*}[!htb]
\centering

\caption{Comparison of translation quality between competition winners by language and our system. The highest scores are highlighted in bold.}
{\normalsize
\resizebox{\linewidth}{!}{%
\begin{tabular}{ll|ccccc|c}
\toprule
\multirow{3}{*}{}         & \multirow{3}{*}{System} & \multicolumn{5}{c|}{Terminology-focused} & \multicolumn{1}{c}{Translation Quality}  \\
                          &                        & Exact-Match & Lemmatized-Match & \multicolumn{2}{c}{Window Overlap} & 1-TERm       &        \\
                          &                        & Accuracy (EMA) & Accuracy (LMA) & (2) & (3) &   Score     & BLEU       \\\midrule
\multirow{5}{*}{EN-FR} & PROMT.soft                    & 0.959 & 0.972 & 0.226 & 0.218 & 0.448 &  41.22 \\
                            & M2M100 small baseline & 0.813 & 0.838 &  0.191 & 0.188 & 0.409 & 35.38 \\
                            & M2M100 small cascaded          & 0.985 & 0.990 &  0.207 & 0.207 & 0.408 & 35.21\\
                            & M2M100 small GBS & 0.989 & 0.993 & 0.158 & 0.160 & 0.207 & 21.86 \\
                            & M2M100 large baseline & 0.880 & 0.899 & 0.231 & 0.225 & 0.442 & 40.15 \\
                            & M2M100 large cascaded          & \textbf{0.991} & \textbf{0.994} & 0.240 & 0.236 & 0.444 & 40.51\\
                            & M2M100 large GBS & 0.994 & 0.996 & 0.185 & 0.183 & 0.309 & 27.76 \\\cmidrule{2-8}
                            & NLLB & 0.903 & 0.922 & \textbf{0.340} & \textbf{0.334} & \textbf{0.568} & \textbf{46.68}\\
\midrule
\multirow{5}{*}{EN-KO}  & KEP                           & 0.790 & 0.792 & \textbf{0.052} & \textbf{0.050} & 0.192 & \textbf{16.32} \\
                        & M2M100 small baseline & 0.577 & 0.581 &  0.027 & 0.026 & 0.132 & 10.11 \\
                        & M2M100 small cascaded              & \textbf{0.997} & \textbf{1.000} & 0.044 & 0.044 & 0.128 & 10.40\\
                        & M2M100 small GBS & 0.977 & 0.978 & 0.028 & 0.030 & 0.0 & 4.99 \\
                        & M2M100 large baseline & 0.631 & 0.632 & 0.029 & 0.029 & 0.149 & 11.59 \\
                        & M2M100 large cascaded              & 0.978 & 0.982 & 0.043 & 0.045 & 0.151 & 12.01\\
                        & M2M100 large GBS & 0.981 & 0.981 & 0.030 & 0.031 & 0.026 & 6.65 \\\cmidrule{2-8}
                        & NLLB & 0.652 & 0.660 & 0.042 & 0.042 & \textbf{0.194} & 11.44\\

\midrule
\multirow{5}{*}{EN-RU} & PROMT.soft                    & 0.862 & 0.913 & 0.176 & 0.174 & 0.315 & \textbf{31.22} \\
                            & M2M100 small baseline & 0.682 & 0.725 & 0.136 & 0.132 & 0.274 & 23.89 \\
                            & M2M100 small cascaded          & 0.916 & 0.934 & 0.150 & 0.151 & 0.273 & 23.73\\
                            & M2M100 small GBS & 0.963 & 0.969 & 0.121 & 0.120 & 0.0 & 12.01 \\
                            & M2M100 large baseline & 0.753 & 0.797 & 0.154 & 0.152 & 0.300 & 29.09 \\
                            & M2M100 large cascaded          & \textbf{0.925} & \textbf{0.950} & 0.176 & 0.177 & 0.404 & 29.22\\
                            & M2M100 large GBS & 0.973 & 0.982 & 0.136 & 0.137 & 0.127 & 16.56 \\\cmidrule{2-8}
                            & NLLB & 0.782 & 0.827 & \textbf{0.217} & 0.211 & \textbf{0.419} & 27.76\\
                            
\midrule
\multirow{5}{*}{CZ-DE} & KEP                            & 0.874 & 0.925 & \textbf{0.321} & \textbf{0.313} & \textbf{0.329} & \textbf{34.46} \\
                            & M2M100 small baseline & 0.786 & 0.829 & 0.229 & 0.224 & 0.254 & 22.46 \\
                            & M2M100 small cascaded          & 0.974 & 0.977 & 0.260 & 0.260 & 0.253 & 22.66\\
                            & M2M100 large baseline & 0.857 & 0.902 & 0.290 & 0.281 & 0.300 & 28.59 \\
                            & M2M100 large cascaded          & \textbf{0.971} & \textbf{0.974} & 0.311 & 0.309 & 0.303 & 28.86\\\cmidrule{2-8}
                            & NLLB & 0.721 & 0.786 & 0.253 & 0.250 & 0.297 & 20.78\\

\bottomrule
\end{tabular}
\label{tab:wmt21_full}

}
}
\end{table*}
\section{WMT21 Full Results}
\vfill

\end{document}